\documentclass{article}

    \PassOptionsToPackage{numbers, compress}{natbib}

    \usepackage[preprint]{neurips_2024}

\usepackage{tabularx}
\usepackage{adjustbox}

\usepackage[utf8]{inputenc} 
\usepackage[T1]{fontenc}    
\usepackage{hyperref}       
\usepackage{url}            
\usepackage{booktabs}       
\usepackage{amsfonts}       
\usepackage{nicefrac}       
\usepackage{microtype}      
\usepackage{xcolor}         
\usepackage{amsmath}
\usepackage{multirow,makecell}
\usepackage{graphicx}

\newtheorem{theorem}{Theorem}
\newtheorem{corollary}{Corollary}

\newtheorem{supptheorem}{Theorem}
\newtheorem{suppcorollary}{Corollary}

\newcommand{\alphabar}{\bar{\alpha}}
\newcommand{\xb}{\mathbf{x}}
\newcommand{\yb}{\mathbf{y}}
\newcommand{\nb}{\mathbf{n}}
\newcommand{\Ib}{\mathbf{I}}
\newcommand{\Ab}{\mathbf{A}}
\newcommand{\Mb}{\mathbf{M}}
\newcommand{\Fb}{\mathbf{F}}
\newcommand{\Hb}{\mathbf{H}}
\newcommand{\Db}{\mathbf{D}}

\newcommand{\Expect}{\mathbb{E}}

\title{Bayesian Conditioned Diffusion Models\\ for Inverse Problems}

\author{%
  Alper~G\"{u}ng\"{o}r\\ 
  Aselsan\\
  Ankara, Turkiye\\
  \texttt{alperg@ee.bilkent.edu.tr} \\
  \And
  Bahri Batuhan Bilecen \\
  Aselsan \\
  Ankara, Turkiye \\
  \texttt{batuhan.bilecen@bilkent.edu.tr} \\
  \AND
  Tolga \c{C}ukur \\
  Department of Electrical and Electronics Engineering, Bilkent University \\
  Ankara, Turkiye \\
  \texttt{cukur@ee.bilkent.edu.tr} \\
}

\begin{document}

\maketitle

\begin{abstract}
Diffusion models have recently been shown to excel in many image reconstruction tasks that involve inverse problems based on a forward measurement operator. A common framework uses task-agnostic unconditional models that are later post-conditioned for reconstruction, an approach that typically suffers from suboptimal task performance. While task-specific conditional models have also been proposed, current methods heuristically inject measured data as a naive input channel that elicits sampling inaccuracies. Here, we address the optimal conditioning of diffusion models for solving challenging inverse problems that arise during image reconstruction. Specifically, we propose a novel Bayesian conditioning technique for diffusion models, BCDM, based on score-functions associated with the conditional distribution of desired images given measured data. We rigorously derive the theory to express and train the conditional score-function. Finally, we show state-of-the-art performance in image dealiasing, deblurring, super-resolution, and inpainting with the proposed technique. 
\end{abstract}

\section{Introduction}


The exceptional performance of unconditional diffusion models (DM) in image generation \cite{ho2020denoising, song2020denoising, xiao2021tackling, dhariwal2021diffusion, feng2023score} has sparked interest in their adoption for image reconstruction from measured data. The typical approach has been to train an unconditional DM for denoising and to condition it by iteratively enforcing data consistency based on a forward measurement operator \cite{dps2022diffusion, dmps2022diffusion, dar2022adaptive, song2023pseudoinverseguided, wang2022zero}. However, this unconditional DM framework yields suboptimal performance for reconstruction tasks since model training is uninformed on the forward operator \cite{dar2022adaptive}. This framework also requires parameter selection during inference to weigh in the forward operator against the DM, which compromises reliability under low signal-to-noise ratio (SNR) regimes due to difficulties in proper parameter tuning \cite{dps2022diffusion, dmps2022diffusion, song2023pseudoinverseguided}.

Recent studies have adopted advanced DM methods to improve image reconstruction tasks. \textbf{Unconditional DMs:}~\cite{xia2023diffir} extracted a transformer-based latent code, on which diffusion sampling was conducted for inference efficiency.~\cite{ddrm2022denoising} used an unconditional DM while running inference via unsupervised posterior sampling based on manifold-constrained gradients.~\cite{bansal2024cold} built unconditional DMs via non-noise image degradations, and synthesized clean images via heuristic conditioning objectives during inference. ~\cite{dps2022diffusion} extended diffusion samplers to handle non-linear inverse problems by approximating conditional posterior sampling.~\cite{dmps2022diffusion} modified the noise-perturbed likelihood score with a closed-form approximation to better denoise low-SNR regions.~\cite{softDiff} incorporated the forward operator in the diffusion process for simultaneous degradations due to measurements and Gaussian noise. \textbf{Conditional DMs:}~\cite{delbracio2023inversion} learned the conditional mean of degraded images given measurements, while avoiding an analytical characterization of the forward operator.~\cite{peng2024improving} proposed a plug-and-play method to approximate the conditional posterior mean for reverse diffusion steps during conditional sampling.  In essence, majority of existing methods aim to improve performance by conditioning samplers of originally unconditional DM priors, or training conditional DMs under incomplete assumptions regarding the forward operator. Yet, approaches that diverge from the true conditional score-function for clean images given measurements are bound to yield suboptimal solutions.


In this study, we tackle the suboptimality in the conditioning procedures previously adopted in the literature, and propose a novel technique for Bayesian conditioning of DMs (BCDM) that reliably addresses linear inverse problems measured under Gaussian noise. First, we demonstrate that post-conditioning of unconditional DMs is inadequate even under relatively simple forward operators. To address this limitation, we show that there exists a tractable optimal combination of noise-perturbed image samples and measured data to enable learning a conditional score-function associated with the true conditional distribution of clean images given measurements. Next, we outline the procedures to train an unrolled neural network with a tractable loss function in order to capture this Bayesian conditional score-function. Finally, we provide analytical derivations of a fast implementation of BCDM for common inverse problems, including accelerated MRI reconstruction (i.e., image dealiasing) as well as natural image deblurring, super-resolution, and inpainting.

The primary contributions of this study are summarized as follows:
\begin{itemize}
    \item We propose a Bayesian formulation to condition DMs for solving image reconstruction problems under Gaussian noise.
    \item We illustrate the discrepancy between the heuristically conditioned versus Bayesian conditional score-functions in inverse problems.
    \item We show that a tractable loss function can be utilized to estimate the Bayesian conditional score-function.
    \item We build BCDM based on an unrolled network architecture for reliable solutions to inverse problems.
    \item We demonstrate BCDM against the state-of-the-art for pervasive tasks, including dealiasing, deblurring, super-resolution, and inpainting.
\end{itemize}

\section{Diffusion Models}

\subsection{Preliminary: Diffusion Models}

DMs are generative models that draw samples from a distribution $q(\xb_0)$ through a temporal Markov process. DMs typically employ hundreds of steps to gradually add noise onto $\xb_0$, producing a noisy sequence $\xb_{1 \cdots T}$ with $\xb_t$ drawn from purely noise distribution $\xb_t \sim \mathcal{N}(0,\Ib)$. The relation between $\xb_{t-1}$ and $\xb_t$ can be defined as:
\begin{align}
    \xb_t = \sqrt{\alpha_t} \xb_{t-1} + \sqrt{1 - \alpha_t} \nb_t
\end{align}
where $\xb_t$ is the noisy image at timestep $t$, $\nb_t \sim \mathcal{N}(0,\Ib)$ is the independent identically distributed noise, and $\alpha_t$ is the noise schedule used for diffusion. Due to the Markovian nature of the forward process, the relation between the reference image distribution $\xb_0 \sim q(\xb_0)$ and $\xb_t$ can also be represented with $\alphabar_t = \Pi_{i=1}^t \alpha_i$:
\begin{align}
    \xb_t = \sqrt{\alphabar_t} \xb_{0} + \sqrt{1 - \alphabar_t} \nb_t
\end{align}
The reverse process can be viewed as a variance-preserving (VP) stochastic differential equation (SDE). Reverse steps are taken by gradually removing noise from $\xb_t$ to draw a sample from $q(\xb_0)$. $\xb_{t-1}$ is estimated using $\xb_t$ by computing the score-function $s_{\theta}(\xb_t, t) \sim \nabla_{\xb_t} \log q(\xb_t)$:
\begin{align}
    \xb_{t-1} = \frac{1}{\sqrt{\alpha_t}} (\xb_t + (1 - \alpha_t) s_{\theta}(\xb_t, t)) + \sqrt{1-\alpha_t} \nb_t \label{eq:samplingChain}
\end{align}
This sampling chain samples from $q(\xb_{t-1}; \xb_t)$, which asymptotically converges onto sampling from the distribution $q(\xb_0)$ after $T$ timesteps. Here, $s_{\theta}(\xb_t, t)$ is a deep learning-based estimate of the score-function estimate, trained via a tractable loss:
\begin{align}
    f_{score} = \Expect_{q(\xb_t,\xb_0)} \Vert s_{\theta}(\xb_t, t) - \nabla_{\xb_t} \log q(\xb_t;\xb_0) \Vert^2
\end{align}
which corresponds exactly to training with the truly desired, albeit intractable loss $\Expect_{q(\xb_t)} \Vert s_{\theta}(\xb_t, t) - \nabla_{\xb_t} \log q(\xb_t) \Vert^2$, as shown in \cite{Vincent2011}. During inference, the trained unconditional DM can be used for random image generation starting from a random noise sample $\xb_T \sim \mathcal{N}\left( 0, \Ib \right)$.

\subsection{Previous Conditioning Methods for DMs}
To enable the solution of inverse problems, an unconditional DM must be conditioned on the forward measurement operator \cite{dps2022diffusion, dmps2022diffusion}. In particular, image reconstruction tasks require one to sample from the true conditional distribution $q(\xb_0; \yb)$ where $\yb = \Ab\xb_0 + \nb_0$. Here, $\Ab$ is the forward operator, $\xb_0$ is the image vector, $\nb_0$ is the measurement noise vector, and $\yb$ is the data vector. To arrive at the true conditional distribution via a trained DM, the sampling chain in Eq.~\eqref{eq:samplingChain} requires the computation of $\nabla_{\xb_t} \log q(\xb_t; \yb)$. Using Bayes rule, 
\begin{align}
   \nabla_{\xb_t} \log q(\xb_t; \yb) = \nabla_{\xb_t} \log \frac{q(\yb; \xb_t) q(\xb_t)}{q(\yb)} = \nabla_{\xb_t} \log q(\yb; \xb_t) +  \nabla_{\xb_t} \log q(\xb_t)
\end{align}
Because the explicit form for $q(\yb; \xb_t)$ is unknown, common approaches attempt to approximate $\nabla_{\xb_t} \log q(\yb; \xb_t)$ with $\nabla_{\xb_0} \log q(\yb; \xb_0)$ for $\xb_0 = \xb_t$ \cite{dps2022diffusion}. For $\nb_0 \sim \mathcal{N}(0, \sigma_0 I)$, the modified score-function for conditional image generation then becomes:
\begin{align}
    s_{\theta,cond}(\xb_t, \yb, t) = s_{\theta}(\xb_t, t) - \frac{1}{\sigma_0^2} \Ab^T\left( \Ab\xb_t - \yb\right)
\end{align}

Unfortunately, the assumptions involved in previous conditioning methods are often inaccurate, resulting in performance losses due to discrepancy between the true conditional distribution and the conditioned score-function of the DM.

\subsection{True Conditional Score-Function}

\begin{corollary}
Here we show that the conditional score-function associated with $q(\xb_0; \yb)$ can be computed using Tweedie's formula. The detailed derivation can be found in Appendix A (Sec. \ref{sec:appendixa}):
\begin{align}
 \nabla_{\xb_t} \log q(\xb_t; \yb) = \frac{1}{1 - \alphabar_t} (\sqrt{\alphabar_t} \Expect[\xb_0; \xb_t, \yb] - \xb_t) \label{eq:bayesianSampling}
\end{align}
\end{corollary}
Eq.~\eqref{eq:bayesianSampling} requires evaluating expectation with respect to $q(\xb_0; \xb_t, \yb)$, which can be expressed using Bayes' rule as:
\begin{align}
 q(\xb_0; \xb_t, \yb) &= \frac{q(\xb_t, \yb; \xb_0) q(\xb_0)}{q(\xb_t, \yb)} \\
&= \frac{q(\xb_t, \yb; \xb_0) q(\xb_0)}{\int q(\xb_t, \yb;\xb_0) q(\xb_0) d\xb_0}
\end{align}
Using independence of $\xb_t$ and $\yb$ given $\xb_0$, 
\begin{align}
q(\xb_0; \xb_t, \yb) &= \frac{q(\xb_t; \xb_0) q(\yb;\xb_0) q(\xb_0)}{\int q(\xb_t; \xb_0) q(\yb;\xb_0) q(\xb_0) d\xb_0} \\
\Expect[\xb_0; \xb_t, \yb] &= \frac{\int \xb_0 q(\xb_t; \xb_0) q(\yb;\xb_0) q(\xb_0) d\xb_0}{\int q(\xb_t; \xb_0) q(\yb;\xb_0) q(\xb_0) d\xb_0} \label{eq:expectation_cond}
\end{align}
In contrast, the unconditional score-function for $q(\xb_t)$ is:
\begin{align}
 \nabla_{\xb_t} \log q(\xb_t) &= \frac{1}{1 - \alphabar_t} (\sqrt{\alphabar_t} \Expect[\xb_0; \xb_t] - \xb_t) \\
&= \frac{1}{1 - \alphabar_t} \Big(\frac{\int \sqrt{\alphabar_t} \xb_0 q(\xb_t; \xb_0) q(\xb_0) d\xb_0}{\int q(\xb_t; \xb_0) q(\xb_0) d\xb_0} - \xb_t \Big)\label{eq:expectation_uncond}
\end{align}

Comparing Eq.~\eqref{eq:expectation_cond} against the expectation term in Eq.~\eqref{eq:expectation_uncond}, it can be clearly observed that a simple affine correction on the trained score-function of an unconditional DM cannot reliably approximate the true conditional score-function. This fundamental discrepancy is due to a lack of an explicit relation between $\xb_t$ and $\yb$, in the absence of $\xb_0$. As multiple distinct samples of $\xb_0$ can be consistent with a given pair ($\xb_t$, $\yb$), the assumption that $q(\yb;\xb_t) \sim q(\yb;\xb_0)$ is incorrect in general. Due to similar reasons, diffusion Schrodinger bridges that express score-functions based on a heuristic linear combination of $\yb$ and $\xb_0$~\cite{liu2023i2sb} can also fail to reliably approximate the true conditional score-function.

\subsection{A Simple Illustration of Discrepancy}

Let us assume that $q(\xb_0)$ is a $d$-dimensional discrete random variable with $N$ equally likely outcomes: $q(\xb_0) = \frac{1}{N} \Sigma_i \delta(\xb_0 - \mu_i)$. Then, the unconditional score-function $\nabla_{\xb_t} \log q(\xb_t)$ is:
\begin{align}
\nabla_{\xb_t} \log q(\xb_t) &= \frac{1}{1 - \alphabar_t} \Big(\frac{\int \xb_0 q(\xb_t; \xb_0) \frac{1}{N} \Sigma_i \delta(\xb_0 - \mu_i) d\xb_0}{\int q(\xb_t; \xb_0) \frac{1}{N} \Sigma_i \delta(\xb_0 - \mu_i) d\xb_0} - \xb_t\Big) \\
 &= \frac{1}{1 - \alphabar_t} \Big(\frac{\Sigma_i \mu_i q(\xb_t; \xb_0 = \mu_i)}{\Sigma_i q(\xb_t; \xb_0 = \mu_i) } - \xb_t\Big)
\end{align}
In contrast, the Bayesian conditioned score-function (BCSF, i.e., true conditional score-function) is;
\begin{align}
\nabla_{\xb_t} \log q(\xb_t; \yb) &= \frac{1}{1 - \alphabar_t} \Big(\frac{\int \xb_0 q(\yb;\xb_0) q(\xb_t; \xb_0) \frac{1}{N} \Sigma_i \delta(\xb_0 - \mu_i) d\xb_0}{\int q(\yb;\xb_0) q(\xb_t; \xb_0) \frac{1}{N} \Sigma_i \delta(\xb_0 - \mu_i) d\xb_0} - \xb_t\Big) \\
 &= \frac{1}{1 - \alphabar_t} \Big(\frac{\Sigma_i \mu_i q(\xb_t; \xb_0 = \mu_i) q(\yb;\xb_0 = \mu_i)}{\Sigma_i q(\xb_t; \xb_0 = \mu_i) q(\yb;\xb_0 = \mu_i) } - \xb_t\Big)
\end{align}

\begin{figure}
    \centering
    \includegraphics[width=0.65\textwidth]{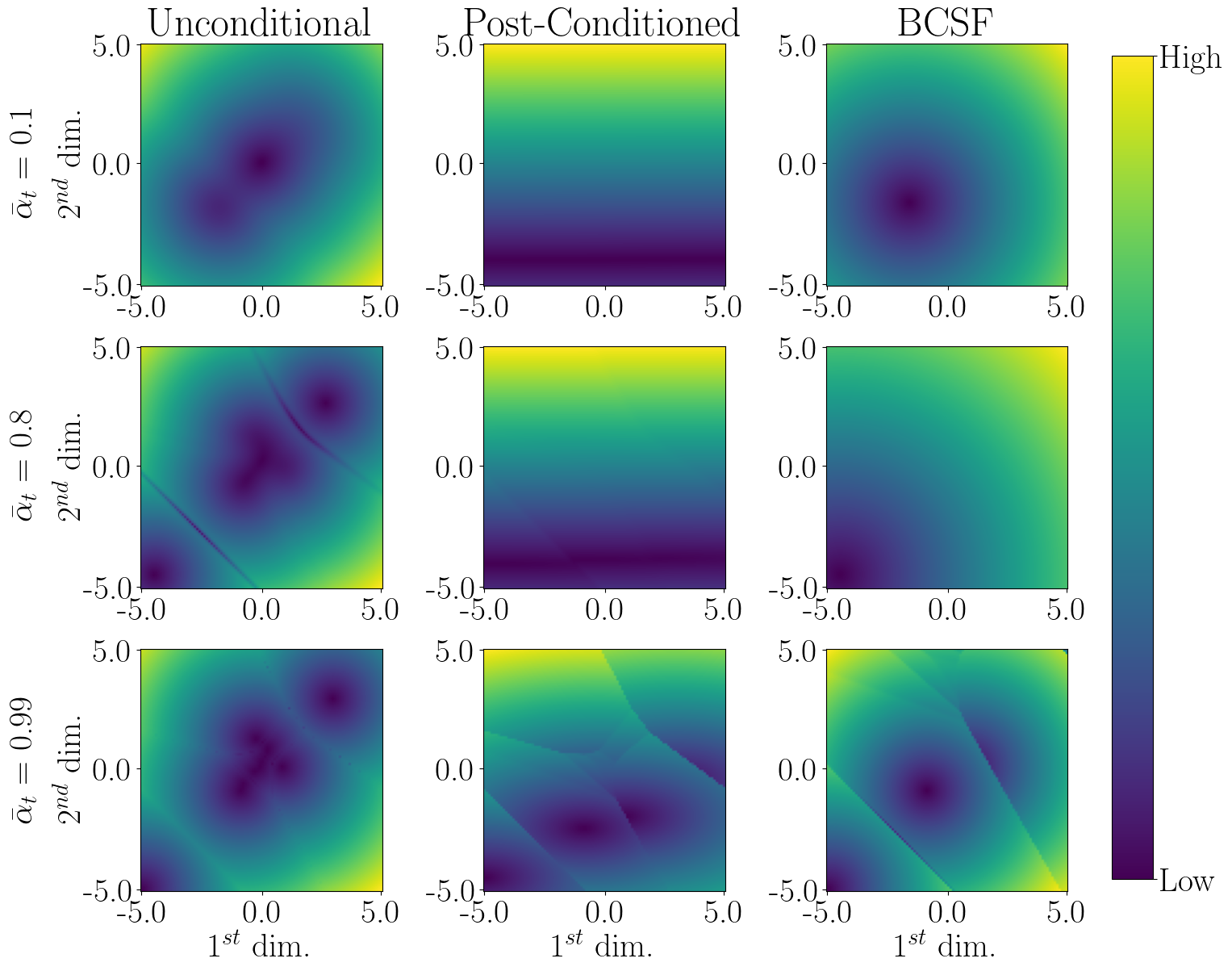}
    \caption{Magnitude of the two-dimensional score-function estimates for unconditional ($\nabla_{\xb_t}\log q(\xb_t)$), post-conditioned ($\nabla_{\xb_t}\log q(\xb_t) + \nabla_{\xb_0} \log q(\yb; \xb_0)$) and Bayesian conditioned score-functions (BCSF, $\nabla_{\xb_t}\log q(\xb_t; \yb)$) are depicted at each column. Discrepancies between post-conditioned and BCSF grow towards lower $\alphabar_t$ values.}
    \label{fig:discrepancySampling}
\end{figure}

Here we provide a simple illustration of the discrepancy between $\nabla_{\xb_t} \log q(\xb_t; \yb)$ (i.e., BCSF) and $\nabla_{\xb_t} \log q(\xb_t) - \frac{1}{\sigma_0^2} \Ab^T\left( \Ab\xb_t - \yb\right)$ (i.e., post-conditioned unconditional score function) as depicted in Fig.~\ref{fig:discrepancySampling}. The forward operator is taken as binary masking $\Ab = [1 \quad 0]$, $\yb = -4$ with $d = 2$ with $N = 12$, and $\sigma_0 = 0.1$. Hence, only a single dimension of the two-dimensional signal $\xb_0$ is measured with additive Gaussian noise. $10$ of the potential outcomes $\mu_i$s are distributed randomly around the origin, while two additional points $\mu_{11} = [-5, -5]$ and $\mu_{12} = [3, 3]$ are placed. The unconditional score-function $\nabla_{\xb_t} \log q(\xb_t)$, the post-conditioned score-function and BCSF $\nabla_{\xb_t} \log q(\xb_t; \yb)$ are shown. It can be observed that the discrepancy between post-conditional and true conditional score-functions grows towards lower $\alphabar_t$, i.e., during initial sampling steps in DMs. 

Prominent manifestations of the abovementioned discrepancy are already apparent in the literature. Aiming to improve sample fidelity, recent studies report that better initialization of the sampler during image generation often improves image quality and/or accelerates inference with existing DMs. For instance,~\cite{softDiff} proposed a momentum sampler to improve sample diversity,~\cite{chung2022ccdf, zhang2024unified} proposed a least-squares or network-driven initialization to improve sampling speed,~\cite{kong2021fast} designed a mapping between continuous diffusion steps and noise levels to shorten reverse process without retraining,~\cite{salimans2021progressive} followed a knowledge distillation approach to progressively improve the sampler, and~\cite{lu2022dpm} proposed a fast diffusion ODE solver via efficient approximations.

\section{Bayesian Conditioning using Correlated Denoisers}

Training of the Bayesian conditioned score-function reflecting the true conditional score-function requires estimation of $\Expect[\xb_0; \xb_t, \yb]$ as evident in Eq.~\eqref{eq:bayesianSampling}. Here, we analytically derive a tractable training objective for BCSF, based on the assumption that acquired measurements are corrupted with additive white Gaussian noise. Note that the additive Gaussian noise scenario is relevant to many practical inverse problems on real-world physical systems \cite{GungorTCI22, chung2022ccdf, dar2022adaptive}. Under this scenario, the forward transition equations at timestep $t$ can be expressed as:
\begin{align}
    \xb_t = \sqrt{\alphabar_t} \xb_0 + \sqrt{1 - \alphabar_t} \nb_t \quad  \yb = \Ab \xb_0 + \sigma_0 \nb_0 \\
    \left[ \begin{array}{c} \xb_t \\ k_t \yb \end{array} \right] = \left[ \begin{array}{c} \sqrt{\alphabar_t} \Ib \\ k_t \Ab  \end{array} \right] \xb_0 + \sqrt{1 - \alphabar_t} \left[ \begin{array}{c} \nb_t \\ \nb_0 \end{array} \right] \label{eq:linearEq}
\end{align}
where $k_t = \frac{\sqrt{1 - \alphabar_t}}{\sigma_0}$. Note that Eq.~\eqref{eq:linearEq} constitutes an over-determined set of linear equations, so $\xb_0$ can be estimated through the pseudo-inverse of $\Ab_t = [ \sqrt{\alphabar_t}\Ib, k_t \Ab^T]^T$, given as:
\begin{align}
    \Ab_t^{\dagger} &= ([ \sqrt{\alphabar_t}\Ib, k_t \Ab^T] [ \sqrt{\alphabar_t}\Ib, k_t \Ab^T]^T)^{-1} [ \sqrt{\alphabar_t}\Ib, k_t \Ab^T] \\
    &= ( \alphabar_t \Ib + k_t^2 \Ab^T\Ab)^{-1} [\sqrt{\alphabar_t}\Ib, k_t \Ab^T]
\end{align}
Multiplying each side of Eq.~\eqref{eq:linearEq} with $\Ab_t^{\dagger}$,
\begin{align}
    ( \alphabar_t \Ib + k_t^2 \Ab^T\Ab)^{-1} (\sqrt{\alphabar_t}\xb_t + k_t^2 \Ab^T \yb) = \xb_0 + \sqrt{1 - \alphabar_t}  ( \alphabar_t \Ib + k_t^2 \Ab^T\Ab)^{-1} (\sqrt{\alphabar_t}\nb_t + k_t \Ab^T \nb_0)
\end{align}
Because $\nb_t$ and $\nb_0$ are independent, $( \alphabar_t \Ib + k_t^2 \Ab^T\Ab)^{-1}(\sqrt{\alphabar_t}\nb_t + k_t \Ab^T \nb_0)$ is a zero-mean multi-variate Gaussian with covariance $( \alphabar_t \Ib + k_t^2 \Ab^T\Ab)^{-1}(k_t^2 \Ab^T \Ab + \alphabar_t \Ib) ( \alphabar_t \Ib + k_t^2 \Ab^T\Ab)^{-1} = (\alphabar_t \Ib + k_t^2 \Ab^T\Ab)^{-1}$. For noise whitening via decorrelation, we precondition the equation by multiplying each side with $(\alphabar_t \Ib + k_t^2 \Ab^T\Ab)^{0.5}$ and derive a decorrelated image variable $\hat{\xb}_t$:
\begin{align}
   \hat{\xb}_t &= ( \alphabar_t \Ib + k_t^2 \Ab^T\Ab)^{-0.5} (\sqrt{\alphabar_t}\xb_t + k_t^2 \Ab^T \yb) \\
   &= ( \alphabar_t \Ib + k_t^2 \Ab^T\Ab)^{0.5} \xb_0 + \sqrt{1 - \alphabar_t} \hat{\nb}_t \\
   &= \Ab_t \xb_0 + \sqrt{1 - \alphabar_t} \hat{\nb}_t
   \label{eq:softerDiffusion}
\end{align}
with $\hat{\nb}_t \sim \mathcal{N}(0, \Ib)$ and $\Ab_t = (\alphabar_t \Ib + k_t^2 \Ab^T\Ab)^{0.5}$. 

\begin{corollary}
$\hat{\xb}_t$ is the optimal combination such that,
\begin{align}
    \Expect[\xb_0; \xb_t, \yb] = \Expect[\xb_0; \hat{\xb}_t].
\end{align}
\end{corollary}
The proof can be found in Appendix B  (Sec. \ref{sec:appendixb}). Hence, the score-function that estimates $\Expect[\xb_0; \hat{\xb}_t]$ can be used in place of the true conditional score-function estimating $\Expect[\xb_0; \xb_t, \yb]$ in Eq.~\eqref{eq:bayesianSampling}:
\begin{align}
     \nabla_{\xb_t} \log q(\xb_t; \yb) = \frac{1}{1 - \alphabar_t} (\sqrt{\alphabar_t} \Expect[\xb_0; \hat{\xb}_t] - \xb_t)\label{eq:bayesianSamplingRe}
\end{align}

\subsection{Score-Function Training for BCDM}

The score-function in Eq. \eqref{eq:bayesianSamplingRe} can be learned via a tractable loss, similar to unconditional DMs \cite{Vincent2011}. Since our objective requires evaluation of $\Expect[\xb_0; \hat{\xb}_t]$ instead of $\nabla_{\hat{\xb}_t} \log q(\hat{\xb}_t)$, we train our model to estimate the expectation of $\xb_0$ instead of incrementally added noise between diffusion timesteps.
\begin{theorem}
The score-function estimator $s_\theta(\hat{\xb}_t, t)$ can be trained using:
\begin{align}
 \mathcal{L}_\text{Bayesian} = \Expect_{q(\hat{\xb}_t, \xb_0)} \Vert s_{\theta}(\hat{\xb}_t, t) - \xb_0 \Vert^2
 \label{eq:BayesianLoss}
\end{align}
\end{theorem}
The proof regarding the suitability of Eq.~\eqref{eq:BayesianLoss} in training the Bayesian score-function can be found in Appendix C (Sec. \ref{sec:appendixc}). Hence, the training procedure for BCDM involves drawing a sample $\xb_0$ from the training set, computing $\hat{\xb}_t$ according to Eq.~\eqref{eq:softerDiffusion}, and predicting $\xb_0$ via the neural network $s_{\theta}(\cdot, t)$ such that the loss in Eq.~\eqref{eq:BayesianLoss} can be computed.

\subsection{Network Architecture of BCDM}

While conventional DMs utilize a denoiser network with time encoding, BCDM requires estimation of $\xb_0$ from an input image that carries degradations due to the forward measurement operator for the reconstruction task. Given the immense success of physics-driven unrolled networks in solving inverse problems, here we leverage an unrolled network with $N_{it}$ iterations of denoising and data-consistency (DC) blocks. We adopt the NCSN++ backbone for the denoising blocks \cite{scoresde}. Receiving an initial input $\xb_{DC,0} = \hat{\xb}_t$ and a denoiser mapping $f_\theta(\cdot, t)$, the network iteratively computes:
\begin{align}
    \xb_{d,i+1} = f_\theta(\xb_{DC,i}, t) 
\end{align}
DC is enforced by solving of the following optimization problem based on the denoiser output $\xb_{d,i+1}$:
\begin{align}
   \xb_{DC,i+1} &= \arg\min_{\xb} \frac{1}{2} \Vert \xb - \xb_{d,i+1}\Vert^2 + \frac{\lambda_t}{2} \Vert \Ab_t \xb - \hat{\xb}_t \Vert^2 \label{eq:dcTerm} \\
   &= (\Ib + \lambda_t \Ab_t^T \Ab_t )^{-1} (\xb_{d,i+1} + \lambda_t \Ab_t^T \hat{\xb}_t)
\end{align}
where $i$ denotes the iteration index ($i = 0 \cdots N_{it} - 1$) and the network output is $\xb_{DC,N_{it}}$. Let us notice that for optimal $\xb = \xb_0$ in Eq.~\eqref{eq:dcTerm}, the distribution $q(\Ab_t \xb_0 - \hat{\xb}_t) \sim \mathcal{N}(0, (1 - \alphabar_t) \Ib)$. Hence, setting $\lambda_t = \lambda/(1 - \alphabar_t)$ ensures constant energy in the DC term across timesteps. 

\subsection{Sampling Procedures for BCDM}

For sampling, the Euler-Maruyama predictor without a corrector was used as in \cite{scoresde}. The sampler follows Eq.~\eqref{eq:samplingChain}. Given $\xb_t$ at timestep $t$, we compute the score-function conditioned on $\yb$ as:
\begin{align}
    \nabla_{\xb_t} \log q(\xb_t; \yb) &= \frac{1}{1 - \alphabar_t} (\sqrt{\alphabar_t} s_\theta( ( \alphabar_t \Ib + k_t^2 \Ab^T\Ab)^{-0.5} (\sqrt{\alphabar_t}\xb_t + k_t^2 \Ab^T \yb) , t)  - \xb_t) \\
     &= \frac{1}{1 - \alphabar_t} (\sqrt{\alphabar_t} s_\theta( \Ab_t^{-1} (\sqrt{\alphabar_t}\xb_t + k_t^2 \Ab^T \yb) , t)  - \xb_t)
\end{align}
Note that the inverse of $\Ab_t = (\alphabar_t \Ib + k_t^2 \Ab^T\Ab)^{0.5}$ can be efficiently computed for a variety of common forward operators, as demonstrated next. The predominant factor in the sampling cost of BCDM pertains to the execution of DC blocks in the unrolled network. Thus, here we derive analytical expressions to efficiently compute $\Ab_t$, $\Ab_t^{-1}$ and DC projections.  

\textbf{MRI Reconstruction (Dealiasing):} MRI reconstructions were demonstrated on coil-combined images from the IXI dataset\footnote{\url{https://brain-development.org/ixi-dataset/}}, under a k-space mask for $8\times$ acceleration. The forward operator is a right-unitary matrix $\Ab = \Mb\Fb$, where $\Mb$ denotes a binary k-space mask (conjugate symmetric for simplicity) and $\Fb$ is the 2D-Fourier transform operator. As such, $\Ab_t=\Fb^T (\alphabar_t \Ib + k_t^2 \Mb^T\Mb)^{0.5} \Fb$ can be implemented via filtering in Fourier domain. The DC block in the unrolled network then performs:
\begin{align}
       \xb_{DC,t} = \Fb^T (\Ib + \lambda_t \alphabar_t \Ib + \lambda_t k_t^2 \Mb^T\Mb)^{-1} \Fb (\xb_{d,t} + \lambda_t \Ab_t^T \hat{\xb}_t)
\end{align}

\textbf{Image Deblurring:} CelebAHQ dataset was used with a Gaussian blur kernel with an std. of $2$ \cite{celebAHQ}. Assuming that the Fourier transform of the blur kernel is a diagonal matrix $\Hb$, the forward operator is defined as $\Ab_t = \Fb^T (\alphabar_t \Ib + k_t^2 |\Hb|^2)^{0.5} \Fb$. The DC projection then becomes:
\begin{align}
       \xb_{DC,t} = \Fb^T (\Ib + \lambda_t \alphabar_t \Ib + \lambda_t k_t^2 |\Hb|^2)^{-1} \Fb (\xb_{d,t} + \lambda_t \Ab_t^T \hat{\xb}_t)
\end{align}

\textbf{Image Super-Resolution (SR):} CelebAHQ dataset was used with a box-downsampling operator \cite{celebAHQ}. The forward operator is $\Ab = \Db$, where $\Db$ is a block-diagonal right-unitary matrix with an SR-factor $r$ with $i$-th block-diagonal $\Ab_{t,i}$ given as:
\begin{align}
    \Ab_{t,i} = (\alphabar_t \Ib + \frac{k_t^2}{r} ones(r) )^{0.5} = (\sqrt{\alphabar_t} \Ib + \frac{\sqrt{k_t^2 + \alphabar_t} - \sqrt{\alphabar_t}}{r} ones(r) )
\end{align}
where $ones(r) \in \mathbb{R}^{r \times r}$ is a matrix filled with ones. The DC projection then becomes:
\begin{align}
       \xb_{DC,t, i} = (\Ib + \lambda_t (\alphabar_t \Ib + \frac{k_t^2}{r} ones(r)) )^{-1} (\xb_{d,t, i} + \lambda_t \Ab_{t, i}^T \hat{\xb}_{t, i}) 
\end{align}
Using Woodbury identity,
\begin{align}
       \xb_{DC,t, i} = \frac{1}{1 + \lambda \alphabar_t } \left(\Ib - \frac{\lambda_t k_t^2}{r(1 + \lambda_t \alphabar_t + \lambda_t k_t^2)} ones(r) \right)  (\xb_{d,t, i} + \lambda_t \Ab_{t, i}^T \hat{\xb}_{t, i})
\end{align}
Similarly, $\Ab_t^{-1}$ required for sampling can be computed using Woodbury identity as:
\begin{align}
       \Ab_{t, i}^{-1} = \frac{1}{\sqrt{\alphabar_t} } \left(\Ib - \frac{\sqrt{k_t^2 + \alphabar_t} - \sqrt{\alphabar_t} }{r\sqrt{k_t^2 + \alphabar_t}} ones(r) \right)
\end{align}

\textbf{Image Inpainting:} CelebAHQ dataset was used with a fixed removal mask located near the image centers \cite{celebAHQ}. The forward operator $\Ab = \Mb$ is a simple binary diagonal matrix, so $\Ab_t = (\alphabar_t \Ib + k_t^2 \Mb^T\Mb)^{0.5}$. The DC projection then becomes:
\begin{align}
       \xb_{DC,t} = (\Ib + \lambda_t \alphabar_t \Ib + \lambda_t k_t^2 \Mb^T\Mb)^{-1} (\xb_{d,t} + \lambda_t \Ab_t^T \hat{\xb}_t)
\end{align}

\begin{figure}[t]
\begin{adjustbox}{width=1\linewidth,center}
 \begin{minipage}{0.01\linewidth}
	\end{minipage}
	\begin{minipage}{0.8\linewidth}
		\centering
  \begin{tabularx}{0.95\textwidth}{*{7}{X} }
     \centering \small Ref & \small\centering LS & \small\centering Unrolled & \small\centering DPS & \small\centering DMPS & \small\centering DI & \small\centering BCDM
  \end{tabularx}
  \end{minipage}
 \end{adjustbox}
\newline
\begin{adjustbox}{width=1\linewidth,center}
 \begin{minipage}{0.01\linewidth}
		\rotatebox{90}{\scriptsize MRI}
	\end{minipage}
	\begin{minipage}{0.8\linewidth}
		\centering
		\includegraphics[scale=0.27]{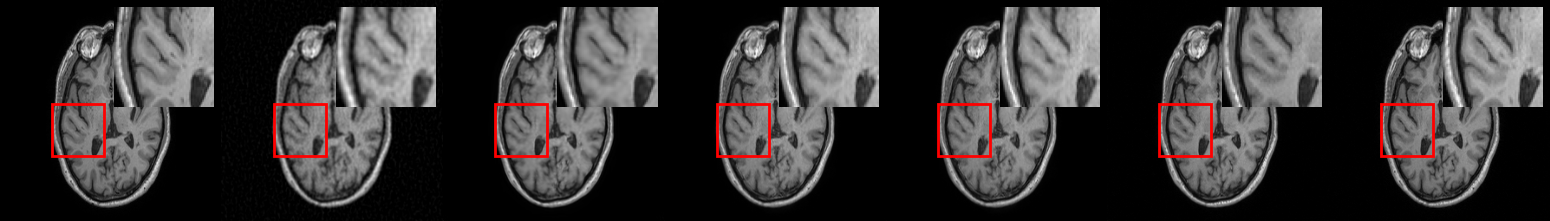}
	\end{minipage}
 \end{adjustbox}
\newline
\begin{adjustbox}{width=1\linewidth,center}
 \begin{minipage}{0.01\linewidth}
		\rotatebox{90}{\scriptsize MRI \\ (Avg.)}
	\end{minipage}
	\begin{minipage}{0.8\linewidth}
		\centering
		\includegraphics[scale=0.27]{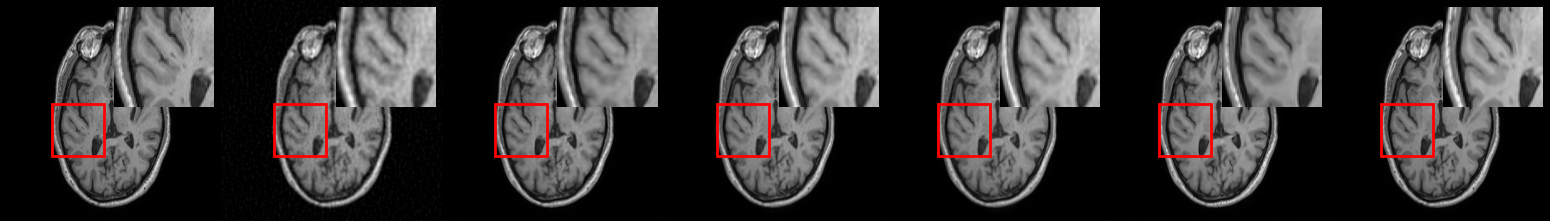}
	\end{minipage}
 \end{adjustbox}
        \newline
\begin{adjustbox}{width=1\linewidth,center}
 \begin{minipage}{0.01\linewidth}
		\rotatebox{90}{\scriptsize Deblur}
	\end{minipage}
	\begin{minipage}{0.8\linewidth}
		\centering
		\includegraphics[scale=0.27]{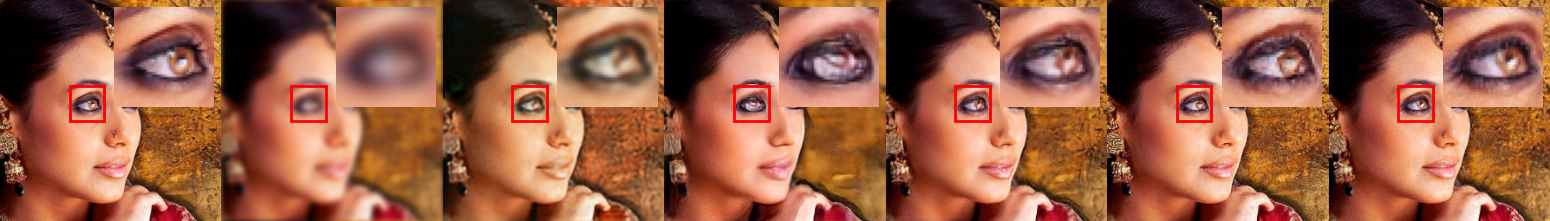}
	\end{minipage}
 \end{adjustbox}
        \newline
\begin{adjustbox}{width=1\linewidth,center}
 \begin{minipage}{0.01\linewidth}
		\rotatebox{90}{\scriptsize Deblur \\ (Avg.)}
	\end{minipage}
	\begin{minipage}{0.8\linewidth}
		\centering
		\includegraphics[scale=0.27]{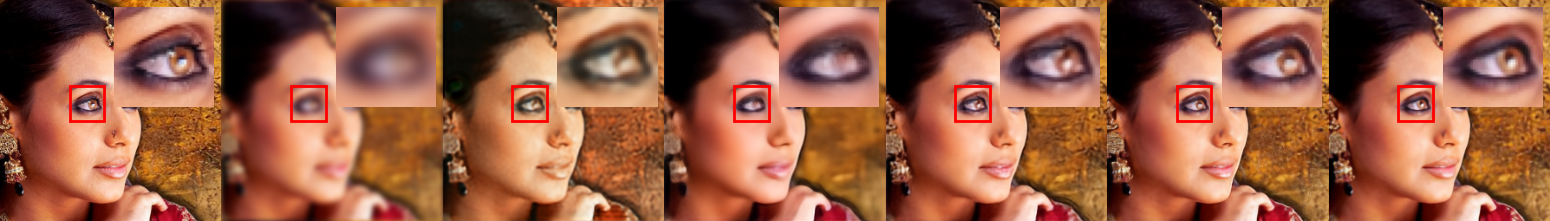}
	\end{minipage}
 \end{adjustbox}
        \newline
\begin{adjustbox}{width=1\linewidth,center}
 \begin{minipage}{0.01\linewidth}
		\rotatebox{90}{\scriptsize SR}
	\end{minipage}
	\begin{minipage}{0.8\linewidth}
		\centering
		\includegraphics[scale=0.27]{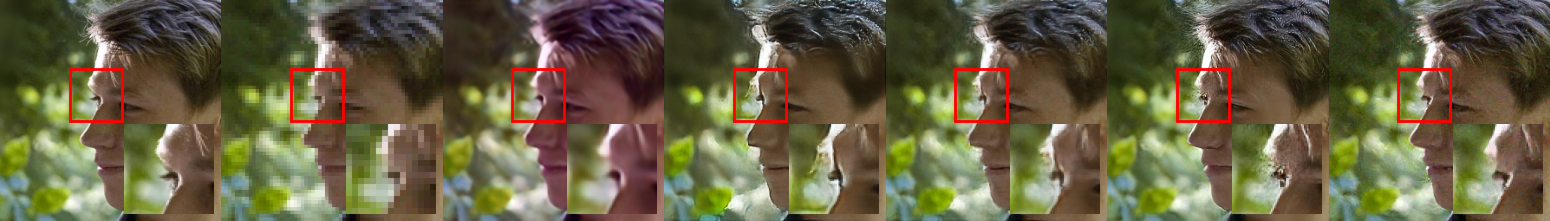}
	\end{minipage}
 \end{adjustbox}
        \newline
\begin{adjustbox}{width=1\linewidth,center}
 \begin{minipage}{0.01\linewidth}
		\rotatebox{90}{\scriptsize 
 SR \\ (Avg.)}
	\end{minipage}
	\begin{minipage}{0.8\linewidth}
		\centering
		\includegraphics[scale=0.27]{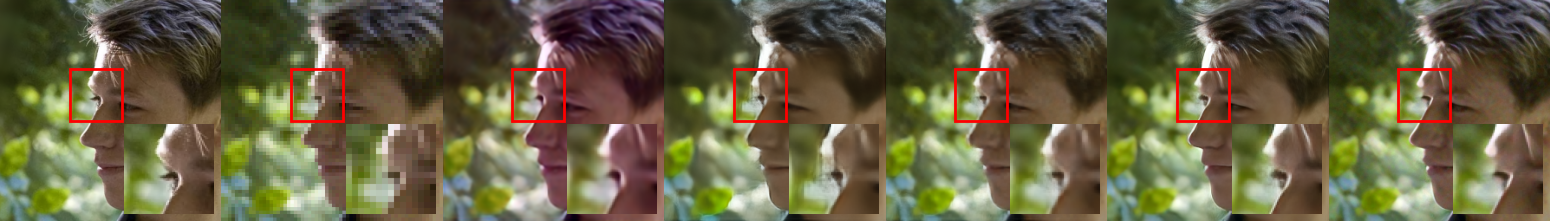}
	\end{minipage}
 \end{adjustbox}
        \newline
\begin{adjustbox}{width=1\linewidth,center}
 \begin{minipage}{0.01\linewidth}
		\rotatebox{90}{\scriptsize Mask}
	\end{minipage}
	\begin{minipage}{0.8\linewidth}
		\centering
		\includegraphics[scale=0.27]{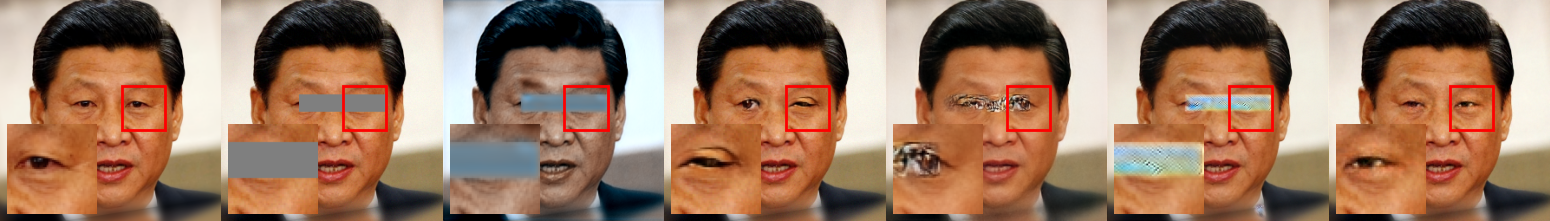}
	\end{minipage}
 \end{adjustbox}
        \newline
\begin{adjustbox}{width=1\linewidth,center}
 \begin{minipage}{0.01\linewidth}
		\rotatebox{90}{\scriptsize Mask \\ (Avg.)}
	\end{minipage}
	\begin{minipage}{0.8\linewidth}
		\centering
		\includegraphics[scale=0.27]{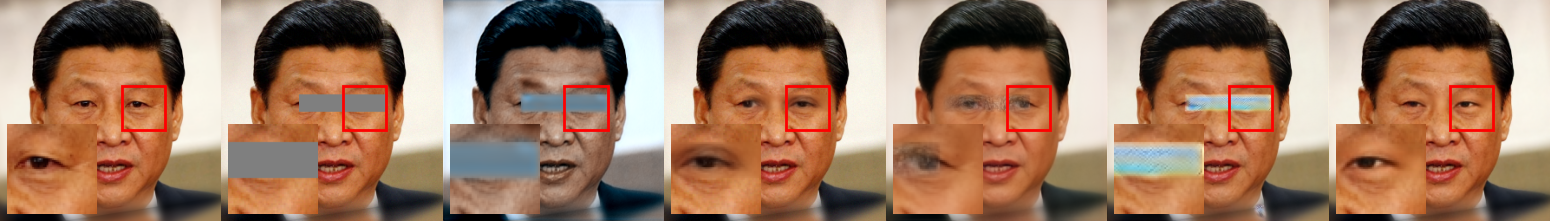}
	\end{minipage}
 \end{adjustbox}
        \newline
	\caption{Representative images for MRI reconstruction (MRI), image deblurring (Deblur), image super-resolution (SR) and image inpainting (Mask) tasks. Columns show reference images (i.e., ground truth), least-squares reconstructions, and images recovered via unrolled, DPS, DMPS, DI and BCDM methods. For each task, the upper row shows a single sample, and the lower row shows average across $10$ samples (Avg.).}
	\label{fig:imagesRecon}
\end{figure}

\section{Results}

\subsection{Experimental Procedures}

We compared the proposed method against a classical unrolled method that does not rely on DM, two post-conditioning methods for unconditional DMs (i.e., DPS \cite{dps2022diffusion} and DMPS \cite{dmps2022diffusion}) and a conditional method with dual-inputs $\xb_t$ and $\yb$ (DI similar to \cite{dhariwal2021diffusion, ozbey2022unsupervised}). While the unrolled method and BCDM utilize iterative networks that cascade denoising and DC blocks, other approaches use non-iterative networks that only perform denoising. The denoising blocks were based on NCSN++ in all methods \cite{scoresde}. Further details are given in the supplementary material (Sec. \ref{sec:experimental}).

For diffusion methods, we averaged reconstructions of $10$ independent images for each input measurement. We reported peak signal-to-noise ratio (pSNR), structural similarity (SSIM) and Frechét Inception Distance (FID) as quantitative performance metrics. Representative images were also visualized for qualitative assessment.

\subsection{MRI Reconstruction}

Quantitative results for MRI reconstruction are summarized in Table~\ref{tab:scores_all}. On average, BCDM achieves (pSNR, SSIM, FID) improvements of (4.9dB, 2.1\%, 40) over post-conditioned DMs (i.e., DPS, DMPS) and (5.7dB, 1.7\%, 23.5) over the conditional DM baseline (i.e., DI). Representative reconstructions in Fig. \ref{fig:imagesRecon} highlight that BCDM yields superior image quality against competing methods with minimal artifacts and acute structural details. While DI is a conditional DM, it cannot alleviate the global aliasing artifacts in accelerated MRI.


\subsection{Image Deblurring}
Quantitative results for image deblurring are summarized in Table~\ref{tab:scores_all}. On average, BCDM achieves (pSNR, SSIM, FID) improvements of (2.5dB, 5.1\%, 19.6) over post-conditioned DMs, and (0.3dB, -0.4\%, 3.8) over DI (albeit DI yields slightly higher SSIM). Note that image blur is a local artifact that DI can more effectively cope with. However, unlike BCDM, DI still fails to provide theoretical guarantees for learning the true conditional score-function. Fig.~\ref{fig:imagesRecon} displays representative images from competing methods. While averaging across independent samples tends to improve image quality for all methods, BCDM shows notably lower artifacts than competing methods in single image instances. 


\begin{table}[t]
    \centering
    \fontsize{7.2pt}{7.2pt}\selectfont
    \setlength\tabcolsep{1.2pt}
    \caption{Quantitative performance of competing methods for various inverse problems. $^*$ depicts significantly different ($p < 0.025$) results compared to BCDM's performance.}
    \begin{tabular}{cccccccccccccccc}
    \toprule
         \multirow{2}{*}{Metric} & \multicolumn{5}{c}{MRI Reconstruction} & \multicolumn{5}{c}{Deblurring} & \multicolumn{5}{c}{Super-resolution} \\
         \cmidrule(lr){2-6}  \cmidrule(lr){7-11}  \cmidrule(lr){12-16}
          & Unrolled & DPS & DMPS & DI & \textbf{BCDM} & Unrolled & DPS & DMPS & DI & \textbf{BCDM} & Unrolled & DPS & DMPS & DI & \textbf{BCDM} \\ 
        \midrule
        
        \multirow{2}{*}{pSNR} 
        & \multirowcell{2}{33.35$^*$ \\ $\pm$3.89} & \multirowcell{2}{34.08$^*$ \\ $\pm$3.26} & \multirowcell{2}{31.55$^*$\\ $\pm$4.22} & \multirowcell{2}{32.08$^*$ \\$\pm$2.58} & \multirowcell{2}{\textbf{37.76} \\ $\pm$\textbf{3.58}} 
        & \multirowcell{2}{29.35$^*$\\$\pm$2.09} & \multirowcell{2}{28.11$^*$\\$\pm$1.72} & \multirowcell{2}{31.11$^*$\\$\pm$2.12} & \multirowcell{2}{31.84$^*$\\$\pm$2.32} & \multirowcell{2}{\textbf{32.12}\\$\pm$\textbf{2.27}}
        & \multirowcell{2}{29.48$^*$\\$\pm$2.20} & \multirowcell{2}{28.29$^*$\\$\pm$1.82} & \multirowcell{2}{29.36$^*$\\$\pm$2.06} & \multirowcell{2}{\textbf{29.95}$^*$\\$\pm$\textbf{2.29}} & \multirowcell{2}{29.57\\$\pm$2.05}
        
        \\ \\ \midrule 
        
        \multirow{2}{*}{SSIM} 
         & \multirowcell{2}{93.1$^*$ \\ $\pm$11.3} & \multirowcell{2}{90.4$^*$ \\ $\pm$11.1} & \multirowcell{2}{\textbf{94.5} \\ $\pm$\textbf{3.9}} & \multirowcell{2}{92.8$^*$ \\ $\pm$7.4} & \multirowcell{2}{94.4 \\ $\pm$9.9}
         & \multirowcell{2}{86.2$^*$ \\ $\pm$3.9} & \multirowcell{2}{81.3$^*$ \\ $\pm$5.1} & \multirowcell{2}{88.6$^*$ \\ $\pm$3.7} & \multirowcell{2}{\textbf{89.7}$^*$ \\ $\pm$\textbf{3.7}} & \multirowcell{2}{89.3 \\ $\pm$3.6} 
         & \multirowcell{2}{85.7$^*$ \\ $\pm$4.5} & \multirowcell{2}{82.5 \\ $\pm$4.8} & \multirowcell{2}{84.7$^*$ \\ $\pm$4.4} & \multirowcell{2}{\textbf{86.4}$^*$ \\ $\pm$\textbf{4.5}} & \multirowcell{2}{82.5 \\ $\pm$3.8} 
        
        \\ \\ \midrule

        FID
        & 65.24 & 48.92 & 55.24 & 35.51 & \textbf{12.04}
        & 45.52 & 41.70 & 19.82 & 14.89 & \textbf{11.13}
        & 44.05 & 33.36 & 39.79 & \textbf{21.62} & 22.76 
        \\
        
    \bottomrule
    \end{tabular}
    \label{tab:scores_all}
\end{table}

\subsection{Image Super-Resolution}
Quantitative results for image super-resolution (SR) are summarized in Table~\ref{tab:scores_all}. In this task, the unrolled method performs more comparably with diffusion methods in terms of pSNR and SSIM, and DI yields the highest performance while BCDM performs competitively with generally second-best metrics among DMs. It is important to note that the common quantitative metrics reported here provide global performance measures that are insufficiently sensitive to detailed image features, thus visual evaluations serve a critical role in comparative assessments, especially for the SR task where high-frequency features are recovered in the absence of high-frequency data. In terms of visual quality, we find that BCDM outperforms competing methods in spatial acuity, particularly near heterogeneous regions containing object boundaries, as exemplified in Fig.~\ref{fig:imagesRecon}.

\subsection{Image Inpainting}
Finally, to assess the effects of domain shifts in the forward operator on reconstruction performance, we conducted an image inpainting experiment. The domain shift was obtained by using larger pixel-removal masks during training than those prescribed during testing. Note that knowledge of the forward operator only influences conditional DMs, while unconditional DMs are not affected (i.e., DPS, DMPS). Quantitative metrics are summarized in Table~\ref{tab:inpatingin}. The unrolled method is a deterministic approach so it faces difficulty in filling in the masked image region. On average, BCDM achieves (pSNR, SSIM, FID) improvements of (2.9dB, 6.2\%, 15.9) over post-conditioned DMs, and (9.7dB, 4.9\%, 269.4) over DI. These results suggest that BCDM is more resilient against domain shifts in the forward operator than other DMs. In terms of visual quality, BCDM substantially outperforms competing methods in terms of accuracy of structural details, as seen in Fig.~\ref{fig:imagesRecon}.

\begin{table}[t]
    \centering
    \fontsize{8pt}{8pt}\selectfont
    \setlength\tabcolsep{1.2pt}
    \caption{Quantitative performance of competing methods for image inpainting. $^*$ depicts significantly different ($p < 0.025$) results compared to BCDM's performance.}
    \label{tab:inpatingin}
\begin{tabular}{cccccc}
    \toprule
         \multirow{2}{*}{Metric} & \multicolumn{5}{c}{Inpainting} \\ \cmidrule(lr){2-6} 
         & Unrolled & DPS & DMPS & DI & \textbf{BCDM} \\ \midrule
pSNR & 11.02$^*\pm$3.04 & 33.72$^*\pm$1.62 & 28.54$^*\pm$1.37 & 24.32$^*\pm$1.08 & \textbf{34.01$\pm$2.27} \\ \midrule 
SSIM & 59.3$^*\pm$11.0 & 94.4$^*\pm$1.1 & 89.2$^*\pm$3.2 & 92.9$^*\pm$3.6 & \textbf{97.5}$\pm$\textbf{2.3} \\ \midrule
FID & 127.37 & 7.31 & 31.67 & 273.05 & \textbf{3.64} \\
\bottomrule
    \end{tabular}
\end{table}

\section{Discussion \& Limitations}

In this study, we introduced an analytically-driven, tractable technique for Bayesian conditioning of DMs for the solution of linear inverse problems with Gaussian measurement noise. While previous methods for DM-based image reconstruction post-condition unconditional DMs under heuristic guidance from data-consistency projections, BCDM directly trains a network to learn the true conditional score-function of clean images given measured data. The proposed technique was comparatively demonstrated against post-conditioning methods as well as a common conditional DM that receives measured data as an added input channel to the recovery network. BCDM generally attained superior quantitative and qualitative performance in key reconstruction tasks, including image dealiasing, deblurring, super-resolution, and inpainting. 

Several extensions could be pursued to enable the adoption of BCDM in a broader range of image reconstruction tasks. Here we focused on Bayesian conditioning under a Gaussian measurement noise model as it is pervasive to many imaging modalities. In principle, the presented derivations can be extended to cover non-Gaussian noise models, although this may require the adoption of non-Gaussian noise corruptions in the diffusion process itself. While our primary derivations for BCDM assume a variance-preserving SDE setup, their extension to variance-exploding SDEs is straightforward. We present the derivations for a VE-SDE variant and a simple demonstration for image denoising in the supplementary material (Sec. \ref{sec:VE-SDE-1}-\ref{sec:VE-SDE-2}). 

An inherent limitation of learning a conditional score-function is that a separate network is trained for each forward operator to ensure optimal performance. While we observed reasonable reliability against domain shifts in the forward operator for BCDM, generalization can be further boosted by multi-task training under diverse forward operators or by employing test-time adaption of the score-function estimator \cite{dar2022adaptive}. 

Another practical limitation pertains to relatively slower inference in BCDM due to the unrolled network compared to DMs that use simpler architectures (see Sec. \ref{sec:experimental} for training-inference times). To accelerate inference, alternate diffusion trajectories could be adopted, for instance, by changing the noise schedule, to maintain similar performance under a fewer number of total timesteps $T$~\cite{liu2022pseudo,watson2021learning,nichol2021improveddiff}. Other alternatives include denoising diffusion implicit model (DDIM)-type samplers \cite{song2020denoising} and hybrid adversarial learning to enable larger step sizes \cite{xiao2021tackling}.

\bibliographystyle{plain} 





\newpage

\section{Supplementary Material}


\subsection{Appendix A.}
\label{sec:appendixa}
\begin{suppcorollary}
The score-function associated with $q(\xb_0; \yb)$ can be computed using Tweedie's formula.
\begin{align}
 \nabla_{\xb_t} \log q(\xb_t; \yb) = \frac{1}{1 - \alphabar_t} (\sqrt{\alphabar_t} \Expect[\xb_0; \xb_t, \yb] - \xb_t). \label{eq:bayesianSampling2}
\end{align}
\end{suppcorollary}

\textbf{Proof:} The score-function for $q(\xb_t; \yb)$ is:
\begin{align}
\nabla_{\xb_t} \log q(\xb_t;\yb) &= \nabla_{\xb_t} \log \frac{q(\xb_t, \yb)}{q(\yb)}, \\
 &= \nabla_{\xb_t} \log q(\xb_t, \yb), \\
 &= \nabla_{\xb_t} \log \int q(\xb_t, \yb;\xb_0) q(\xb_0) d\xb_0.
\end{align}
Using independence of $\xb_t$ and $\yb$ given $\xb_0$;
\begin{align}
\nabla_{\xb_t} \log q(\xb_t;\yb) &= \nabla_{\xb_t} \log \int q(\xb_t; \xb_0) q(\yb;\xb_0) q(\xb_0) d\xb_0, \\
 &= \frac{\nabla_{\xb_t} \int q(\xb_t; \xb_0) q(\yb;\xb_0) q(\xb_0) d\xb_0}{\int q(\xb_t; \xb_0) q(\yb;\xb_0) q(\xb_0) d\xb_0}, \\
 &= \frac{\int \nabla_{\xb_t} q(\xb_t; \xb_0) q(\yb;\xb_0) q(\xb_0) d\xb_0}{\int q(\xb_t; \xb_0) q(\yb;\xb_0) q(\xb_0) d\xb_0}.
\end{align}
Here, since $q(\xb_t;\xb_0)$ is of exponential distribution family; 
\begin{align}
    \nabla_{\xb_t} q(\xb_t; \xb_0) = -\frac{1}{1 - \alphabar_t} (\xb_t - \sqrt{\alphabar_t} \xb_0 ) q(\xb_t; \xb_0).
\end{align}
Hence,
\begin{align}
    &\nabla_{\xb_t} \log q(\xb_t;\yb) = -\frac{1}{1 - \alphabar_t} \frac{\int (\xb_t - \sqrt{\alphabar_t} \xb_0 ) q(\xb_t; \xb_0) q(\yb;\xb_0) q(\xb_0) d\xb_0}{\int q(\xb_t; \xb_0) q(\yb;\xb_0) q(\xb_0) d\xb_0}, \\
     &= -\frac{1}{1 - \alphabar_t} \left( \xb_t \frac{\int q(\xb_t; \xb_0) q(\yb;\xb_0) q(\xb_0) d\xb_0}{\int q(\xb_t; \xb_0) q(\yb;\xb_0) q(\xb_0) d\xb_0} -  \frac{\int \sqrt{\alphabar_t} \xb_0 q(\xb_t; \xb_0) q(\yb;\xb_0) q(\xb_0) d\xb_0}{\int q(\xb_t; \xb_0) q(\yb;\xb_0) q(\xb_0) d\xb_0} \right), \\
     &= -\frac{1}{1 - \alphabar_t} \left( \xb_t -  \frac{\int \sqrt{\alphabar_t} \xb_0 q(\xb_t,y, \xb_0) d\xb_0}{q(\xb_t,\yb)} \right), \\
     &= \frac{1}{1 - \alphabar_t} \left( \int \sqrt{\alphabar_t} \xb_0 q(\xb_0; \xb_t, \yb) d\xb_0 - \xb_t \right), \\
     &= \frac{1}{1 - \alphabar_t} \left( \sqrt{\alphabar_t} \Expect[\xb_0; \xb_t, \yb] - \xb_t \right) \square
\end{align}

\subsection{Appendix B.}
\label{sec:appendixb}
\begin{suppcorollary}
$\hat{\xb}_t$ is the optimal combination such that,
\begin{align}
    \Expect[\xb_0; \xb_t, \yb] = \Expect[\xb_0; \hat{\xb}_t].
\end{align}
\end{suppcorollary}
\textbf{Proof:} For measurement processes with Gaussian noise corruption, $d$-dimensional $\xb_0$ and $m$-dimensional $\yb = \Ab\xb_0 + \nb_0$, the expectation for Bayesian conditioning is:
\begin{align}
    \Expect[&\xb_0; \xb_t, \yb] = \int \xb_0 q(\xb_0; \xb_t, \yb) d\xb_0, \\
     &= \int \xb_0 \frac{q(\xb_t, \yb;\xb_0) q(\xb_0)}{q(\xb_t, \yb)} d\xb_0, \\
     &= \frac{\int \xb_0 q(\xb_t; \xb_0) q(\yb;\xb_0) q(\xb_0) d\xb_0}{\int q(\xb_t; \xb_0) q(\yb;\xb_0) q(\xb_0) d\xb_0}, \\
     =& \frac{\int \xb_0 \frac{1}{(2 \pi (1-\alphabar_t) )^{d/2} } \exp \{ -\frac{1}{2(1-\alphabar_t)} \Vert \xb_t - \sqrt{\alphabar}_t \xb_0 \Vert^2  \} \frac{1}{(2 \pi \sigma_0^2 )^{m/2} } \exp \{ -\frac{1}{2\sigma_0^2} \Vert \yb - \Ab \xb_0 \Vert^2  \} q(\xb_0) d\xb_0}{\int \frac{1}{(2 \pi (1-\alphabar_t) )^{d/2} } \exp \{ -\frac{1}{2(1-\alphabar_t)} \Vert \xb_t - \sqrt{\alphabar}_t \xb_0 \Vert^2  \} \frac{1}{(2 \pi \sigma_0^2 )^{m/2} } \exp \{ \frac{1}{2\sigma_0^2} \Vert \yb - \Ab \xb_0 \Vert^2  \} q(\xb_0) d\xb_0}, \\
     &= \frac{\int \xb_0 \exp \{ -\frac{1}{2(1-\alphabar_t)} \Vert \xb_t - \sqrt{\alphabar}_t \xb_0 \Vert^2 -\frac{1}{2\sigma_0^2} \Vert \yb - \Ab \xb_0 \Vert^2  \} q(\xb_0) d\xb_0}{\int \exp \{ -\frac{1}{2(1-\alphabar_t)} \Vert \xb_t - \sqrt{\alphabar}_t \xb_0 \Vert^2 -\frac{1}{2\sigma_0^2} \Vert \yb - \Ab \xb_0 \Vert^2  \} q(\xb_0) d\xb_0}, \\
     &= \frac{\int \xb_0 \exp \{ -\frac{1}{2(1-\alphabar_t)} \left( \alphabar_t \xb_0^T \xb_0 - 2 \sqrt{\alphabar_t} \xb_0^T \xb_t - 2 k_t^2 \xb_0^T \Ab^T \yb + k_t^2 \xb_0^T \Ab^T \Ab \xb_0 \right)\} q(\xb_0) d\xb_0}{\int \exp \{ -\frac{1}{2(1-\alphabar_t)} \left( \alphabar_t \xb_0^T \xb_0 - 2 \sqrt{\alphabar_t} \xb_0^T \xb_t - 2 k_t^2 \xb_0^T \Ab^T \yb + k_t^2 \xb_0^T \Ab^T \Ab \xb_0 \right)\} q(\xb_0) d\xb_0}, \\
     &= \frac{\int \xb_0 \exp \{ -\frac{1}{2(1-\alphabar_t)} \left( \xb_0^T (\alphabar_t\Ib + k_t^2 \Ab^T \Ab ) \xb_0 - 2 \xb_0^T (\sqrt{\alphabar_t} \xb_t + k_t^2 \Ab^T \yb)  \right)\} q(\xb_0) d\xb_0}{\int \exp \{ -\frac{1}{2(1-\alphabar_t)} \left( \xb_0^T (\alphabar_t\Ib + k_t^2 \Ab^T \Ab ) \xb_0 - 2 \xb_0^T (\sqrt{\alphabar_t} \xb_t + k_t^2 \Ab^T \yb)  \right)\} q(\xb_0) d\xb_0}. \label{eq:appendixEx0givenxty}
\end{align}

Since $q(\hat{\xb}_t; \xb_0) \sim \mathcal{N}(( \alphabar_t \Ib + k_t^2 \Ab^T\Ab)^{0.5} \xb_0, (1 - \alphabar_t) \Ib)$:
\begin{align}
    \Expect[&\xb_0; \hat{\xb}_t] = \int \xb_0 q(\xb_0; \hat{\xb}_t) d\xb_0, \\
     &= \int \xb_0 \frac{q(\hat{\xb}_t; \xb_0) q(\xb_0)}{q(\hat{\xb}_t)} d\xb_0, \\
     &= \frac{\int \xb_0 q(\hat{\xb}_t; \xb_0) q(\xb_0) d\xb_0}{\int q(\hat{\xb}_t; \xb_0) q(\xb_0) d\xb_0}, \\
     &= \frac{\int \xb_0 \frac{1}{(2 \pi (1-\alphabar_t) )^{d/2} } \exp \{ -\frac{1}{2(1-\alphabar_t)} \Vert \hat{\xb}_t - ( \alphabar_t \Ib + k_t^2 \Ab^T\Ab)^{0.5} \xb_0 \Vert^2  \} q(\xb_0) d\xb_0}{\int \frac{1}{(2 \pi (1-\alphabar_t) )^{d/2} } \exp \{ -\frac{1}{2(1-\alphabar_t)} \Vert \hat{\xb}_t - ( \alphabar_t \Ib + k_t^2 \Ab^T\Ab)^{0.5} \xb_0 \Vert^2  \} q(\xb_0) d\xb_0}, \\
     &= \frac{\int \xb_0 \exp \{ -\frac{1}{2(1-\alphabar_t)} \Vert \hat{\xb}_t - ( \alphabar_t \Ib + k_t^2 \Ab^T\Ab)^{0.5} \xb_0 \Vert^2  \} q(\xb_0) d\xb_0}{\int \exp \{ -\frac{1}{2(1-\alphabar_t)} \Vert \hat{\xb}_t - ( \alphabar_t \Ib + k_t^2 \Ab^T\Ab)^{0.5} \xb_0 \Vert^2  \} q(\xb_0) d\xb_0},
\end{align}
Since $\hat{\xb}_t = ( \alphabar_t \Ib + k_t^2 \Ab^T\Ab)^{-0.5} (\sqrt{\alphabar_t}\xb_t + k_t^2 \Ab^T \yb)$, 
\begin{align}
    \Expect[&\xb_0; \hat{\xb}_t] = \nonumber \\
    &\frac{\int \xb_0 \exp \{ -\frac{1}{2(1-\alphabar_t)} \Vert ( \alphabar_t \Ib + k_t^2 \Ab^T\Ab)^{-0.5} (\sqrt{\alphabar_t}\xb_t + k_t^2 \Ab^T \yb) - ( \alphabar_t \Ib + k_t^2 \Ab^T\Ab)^{0.5} \xb_0 \Vert^2  \} q(\xb_0) d\xb_0}{\int \exp \{ -\frac{1}{2(1-\alphabar_t)} \Vert ( \alphabar_t \Ib + k_t^2 \Ab^T\Ab)^{-0.5} (\sqrt{\alphabar_t}\xb_t + k_t^2 \Ab^T \yb) - ( \alphabar_t \Ib + k_t^2 \Ab^T\Ab)^{0.5} \xb_0 \Vert^2  \} q(\xb_0) d\xb_0}, \\
    =& \frac{\int \xb_0 \exp \{ -\frac{1}{2(1-\alphabar_t)} \Vert ( \alphabar_t \Ib + k_t^2 \Ab^T\Ab)^{-0.5} (\sqrt{\alphabar_t}\xb_t + k_t^2 \Ab^T \yb) - ( \alphabar_t \Ib + k_t^2 \Ab^T\Ab)^{0.5} \xb_0 \Vert^2  \} q(\xb_0) d\xb_0}{\int \exp \{ -\frac{1}{2(1-\alphabar_t)} \Vert ( \alphabar_t \Ib + k_t^2 \Ab^T\Ab)^{-0.5} (\sqrt{\alphabar_t}\xb_t + k_t^2 \Ab^T \yb) - ( \alphabar_t \Ib + k_t^2 \Ab^T\Ab)^{0.5} \xb_0 \Vert^2  \} q(\xb_0) d\xb_0}, \\
    &= \frac{\int \xb_0 \exp \{ -\frac{1}{2(1-\alphabar_t)} \left( \xb_0^T ( \alphabar_t \Ib + k_t^2 \Ab^T\Ab) \xb_0 - 2 \xb_0^T (\sqrt{\alphabar_t}\xb_t + k_t^2 \Ab^T \yb) \right) \} q(\xb_0) d\xb_0}{\int \exp \{ -\frac{1}{2(1-\alphabar_t)} \left( \xb_0^T ( \alphabar_t \Ib + k_t^2 \Ab^T\Ab) \xb_0 -2 \xb_0^T (\sqrt{\alphabar_t}\xb_t + k_t^2 \Ab^T \yb) \right) \} q(\xb_0) d\xb_0}. \label{eq:appendixEx0givenxhatt}
\end{align}
Since Eq.~\eqref{eq:appendixEx0givenxty} and Eq.~\ref{eq:appendixEx0givenxhatt} are equal, $\Expect[\xb_0; \hat{\xb}_t] = \Expect[\xb_0; \xb_t, \yb]$. Hence, correlated noise denoising with the forward model $(\alphabar_t \Ib + k_t^2 \Ab^T\Ab)^{0.5}$ exactly corresponds to computing conditional expectation. $\square$

\subsection{Appendix C.}
\label{sec:appendixc}
\begin{supptheorem}
The score-function estimator $s_\theta(\hat{\xb}_t, t)$ can be trained using:
\begin{align}
 \mathcal{L}_\text{Bayesian} = \Expect_{q(\hat{\xb}_t, \xb_0)} \Vert s_{\theta}(\hat{\xb}_t, t) - \xb_0 \Vert^2. \label{eq:BayesianLossApp}
\end{align}
\end{supptheorem}

\textbf{Proof:} To train on $\Expect[\xb_0; \hat{\xb}_t]$, we first use the Tweedie's formula to estimate $\Expect[\xb_0; \hat{\xb}_t]$ using the score-function:
\begin{align}
 \nabla_{\hat{\xb}_t} \log q(\hat{\xb}_t) &= \frac{1}{1 - \alphabar_t} ( \Expect[\Ab_t \xb_0; \hat{\xb}_t] - \hat{\xb}_t), \\
 \nabla_{\hat{\xb}_t} \log q(\hat{\xb}_t) &= \frac{1}{1 - \alphabar_t} ( \Ab_t \Expect[\xb_0; \hat{\xb}_t] - \hat{\xb}_t), \\
 \Expect[\xb_0; \hat{\xb}_t] &= \Ab_t^{-1} \left(\hat{\xb}_t - (1 - \alphabar_t) \nabla_{\hat{\xb}_t} \log q(\hat{\xb}_t) \right).
\end{align}
Hence, if the network $s_\theta(\hat{\xb}_t, t)$ estimates $\Expect[\xb_0; \hat{\xb}_t]$, then it needs to be trained with the loss:
\begin{align}
    L_{\xb_0} = \Expect_{q(\hat{\xb}_t)} \Vert s_{\theta}(\hat{\xb}_t, t) - \Ab_t^{-1} \left(\hat{\xb}_t - (1 - \alphabar_t) \nabla_{\hat{\xb}_t} \log q(\hat{\xb}_t) \right) \Vert^2.
\end{align}
Following similar arguments to \cite{Vincent2011} and \cite{softDiff}, the loss consists of two terms that depends on $\theta$: 
\begin{align}
&\arg\min_{\theta} \int \left(\Vert s_{\theta}(\hat{\xb}_t, t) \Vert^2 - s_{\theta}(\hat{\xb}_t, t)^T \Ab_t^{-1} \left(\hat{\xb}_t - (1 - \alphabar_t) \nabla_{\hat{\xb}_t} \log q(\hat{\xb}_t) \right) \right) q(\hat{\xb}_t) d\hat{\xb}_t.    
\end{align}
Focusing on the first term;
\begin{align}
    &= \int \Vert s_{\theta}(\hat{\xb}_t, t) \Vert^2 q(\hat{\xb}_t) d\hat{\xb}_t, \\
    &= \iint \Vert s_{\theta}(\hat{\xb}_t, t) \Vert^2 q(\hat{\xb}_t, \xb_0) d\hat{\xb}_t d\xb_0, \\
    &= \Expect_{q(\hat{\xb}_t, \xb_0)} \left[\Vert s_{\theta}(\hat{\xb}_t, t) \Vert^2 \right].
\end{align}
Focusing on the second term;
\begin{align}
    =&\int s_{\theta}(\hat{\xb}_t, t)^T \Ab_t^{-1} \left(\hat{\xb}_t - (1 - \alphabar_t) \nabla_{\hat{\xb}_t} \log q(\hat{\xb}_t) \right) q(\hat{\xb}_t) d\hat{\xb}_t, \\
    =&\int s_{\theta}(\hat{\xb}_t, t)^T \Ab_t^{-1} \left(\hat{\xb}_t - (1 - \alphabar_t) \frac{\nabla_{\hat{\xb}_t} q(\hat{\xb}_t)}{q(\hat{\xb}_t)} \right) q(\hat{\xb}_t) d\hat{\xb}_t, \\
    =&\int \left( s_{\theta}(\hat{\xb}_t, t)^T \Ab_t^{-1} \hat{\xb}_t q(\hat{\xb}_t) -(1 - \alphabar_t) s_{\theta}(\hat{\xb}_t, t)^T \Ab_t^{-1} \nabla_{\hat{\xb}_t} q(\hat{\xb}_t) \right) d\hat{\xb}_t, \\
    =&\int \left( s_{\theta}(\hat{\xb}_t, t)^T \Ab_t^{-1} \hat{\xb}_t \int q(\hat{\xb}_t, \xb_0) d\xb_0 - (1 - \alphabar_t) s_{\theta}(\hat{\xb}_t, t)^T \Ab_t^{-1} \nabla_{\hat{\xb}_t} \int q(\hat{\xb}_t; \xb_0) q(\xb_0) d\xb_0\right) d\hat{\xb}_t, \\
    =&\iint \left( s_{\theta}(\hat{\xb}_t, t)^T \Ab_t^{-1} \hat{\xb}_t q(\hat{\xb}_t, \xb_0)  - (1 - \alphabar_t) q(\xb_0) s_{\theta}(\hat{\xb}_t, t)^T \Ab_t^{-1} \nabla_{\hat{\xb}_t} q(\hat{\xb}_t; \xb_0) \right) d\hat{\xb}_td\xb_0, \\
    =&\iint \left( s_{\theta}(\hat{\xb}_t, t)^T \Ab_t^{-1} \hat{\xb}_t  - (1 - \alphabar_t)  s_{\theta}(\hat{\xb}_t, t)^T \Ab_t^{-1} \nabla_{\hat{\xb}_t} \log q(\hat{\xb}_t; \xb_0) \right) q(\hat{\xb}_t, \xb_0) d\hat{\xb}_td\xb_0, \\
    =& \Expect_{q(\hat{\xb}_t, \xb_0)} [ s_{\theta}(\hat{\xb}_t, t)^T \Ab_t^{-1} \left( \hat{\xb}_t  - (1 - \alphabar_t) \nabla_{\hat{\xb}_t} \log q(\hat{\xb}_t; \xb_0) \right)].
\end{align}
Further simplifying,
\begin{align}
    =& \Expect_{q(\hat{\xb}_t, \xb_0)} [ s_{\theta}(\hat{\xb}_t, t)^T  \Ab_t^{-1} \left(\hat{\xb}_t  - (\hat{\xb}_t - \Ab_t \xb_0) \right)], \\
    =& \Expect_{q(\hat{\xb}_t, \xb_0)} [ s_{\theta}(\hat{\xb}_t, t)^T  \Ab_t^{-1} \left(\Ab_t \xb_0 \right)], \\
    =& \Expect_{q(\hat{\xb}_t, \xb_0)} [ s_{\theta}(\hat{\xb}_t, t)^T  \xb_0].
\end{align}
Since adding non-$\theta$ dependent terms does not change the overall loss, following a similar argument to \cite{Vincent2011}, one can conduct training based on the following objective:
\begin{align}
    \Expect_{q(\hat{\xb}_t, \xb_0)} \Vert s_{\theta}(\hat{\xb}_t, t) - \xb_0 \Vert^2 \quad \square
\end{align}








\subsection{Details on Experimental Procedures}
\label{sec:experimental}
Two public datasets were used to demonstrate the proposed method. For MRI reconstruction, the IXI dataset\footnote{\url{https://brain-development.org/ixi-dataset/}} with coil-combined images was used with the train, validation, and test split as in \cite{dar2022adaptive}. Since the network architectures were mostly utilized from previous literature that optimized the architecture, validation sets were not utilized. The training and test sets included $2400$ and $1620$ images. For the remaining tasks, the CelebAHQ dataset was used with default train, validation, and test split with $27000$ and $3000$ images in the train and test sets, respectively.

All experiments were conducted using the PyTorch library on a server equipped with Tesla V100 GPUs and Intel Xeon CPUs. Training was performed for $500$ epochs for each network using 2 GPUs. For training, the ADAM optimizer with a learning rate of $2e-4$ and $(\beta_1, \beta_2) = (0.9, 0.999)$ was used. Training times are given in Table~\ref{tab:trainingTimes}. Inference times are summarized in Table~\ref{tab:inferenceTimes}. Note that the unrolled architecture (i.e., iterative utilization of the denoiser) increases inference times $4$-fold compared to DI. Hence, optimizing the network architecture can help lower the computational costs of the proposed method.

\textbf{Common Network Parameters:} The NCSN++ backbone was utilized as the denoiser network in all experiments \cite{scoresde}. The parameter set was based on variance-preserving CelebAHQ configuration with a few differences. The number of channel multipliers were selected as $[1, 1, 2, 2, 4, 4]$ , and embedding type was selected as Fourier. Specific differences are described for each competing method below. The number of timesteps for diffusion was set to $T = 1000$ for all experiments.

\textbf{Unrolled Method:} The data consistency in the unrolled approach was implemented as in the proposed Bayesian approach. The effect of the temporal encoding was removed by setting a number of temporal encoding channels $n_f = 4$, and constant $t = 0$ input was utilized. Hyper-parameters were selected as $\lambda = 0.01$, and a number of iterations were selected as $10$. This network was trained for $200$ epochs using a single GPU.

\textbf{Post-conditioned DMs:} The number of temporal encoding channels were selected as $n_f = 128$ for the MRI reconstruction experiment and $n_f = 64$ for CelebAHQ experiments. Post-conditioning using DPS and DMPS were implemented as in their respective papers.

\textbf{DI:} An architecture similar to unconditional diffusion was utilized for the dual-input with $n_f = 32$ with Fourier-type embeddings. 

\textbf{BCDM:} The number of temporal channel encodings was selected as $n_f = 32$ with Fourier-type embeddings. The number of iterations was set to $N_{it}4$ with $\lambda = 0.01$. Based on observations in the initial stages of the study, we fractioned the diffusion process to train to separate networks, where the first handled $50 < T < 1000$ and the second handled $T < 50$.

\begin{table}[]
    \centering
    \caption{Training times in hours for competing methods under various tasks. Post-conditioning refers to methods that condition pre-trained unconditional DMs on the forward operator during inference (i.e., DPS, DMPS).}
    \setlength\tabcolsep{1.2pt}
    \begin{tabular}{rcccc}
     \toprule
    & Unrolled & Post-cond. & DI & \textbf{BCDM} \\ \midrule
        MRI Reconstruction & 19 & 43 & 4 & 38 \\ 
        Deblurring & 220 & 83 & 44 & 207 \\ 
        Inpainting & 216 & 83 & 19 & 213 \\ 
        Super-Resolution & 217 & 83 & 12 & 208 \\
        \bottomrule
    \end{tabular}
    \label{tab:trainingTimes}
\end{table}

\begin{table}[]
    \centering
    \caption{Inference times in seconds for competing methods under various tasks.}
    \setlength\tabcolsep{2.0pt}
    \begin{tabular}{rccccc}
     \toprule
    & Unrolled & DPS & DMPS & DI & \textbf{BCDM} \\ \midrule
        MRI Reconstruction & 0.3 & 377.8 & 540 & 64 & 255.6 \\ 
        Deblurring & 0.24 & 135 & 135 & 66.6 & 264.6 \\ 
        Inpainting & 0.3 & 134.4 & 134.4 & 65.4 & 255.4 \\ 
        Super-Resolution & 0.24 & 133.2 & 294 & 65.4 & 256.2 \\ \bottomrule
    \end{tabular}
    \label{tab:inferenceTimes}
\end{table}




\subsection{Bayesian Conditioning of Variance Exploding SDEs}
\label{sec:VE-SDE-1}
A similar derivation can also be shown for variance-exploding SDEs. As in VP-SDEs, the training of Bayesian score-functions requires the estimation of $\Expect[\xb_0; \xb_t, \yb]$. The set of equations at time $t$ is:
\begin{align}
    \xb_t = \xb_0 + \sigma_t \nb_t,  \yb = \Ab \xb_0 + \sigma_0 \nb_0, \\
    \left[ \begin{array}{c} \xb_t \\ v_t \yb \end{array} \right] = \left[ \begin{array}{c} \Ib \\ v_t \Ab  \end{array} \right] \xb_0 + \sigma_t \left[ \begin{array}{c} \nb_t \\ \nb_0 \end{array} \right] , \label{eq:linearEqVE}
\end{align}
where $v_t = \frac{\sigma_t}{\sigma_0}$. Eq.~\eqref{eq:linearEqVE} is a set of over-determined linear set of equations. Then, the pseudo-inverse of $\hat{\Ab}_t = [\Ib, v_t \Ab^T]^T$ is:
\begin{align}
    \hat{\Ab}_t^{\dagger} &= ([ \Ib, v_t \Ab^T] [ \Ib, v_t \Ab^T]^T)^{-1} [ \Ib, v_t \Ab^T], \\
    \hat{\Ab}_t^{\dagger} &= ( \Ib + v_t^2 \Ab^T\Ab)^{-1} [\Ib, v_t \Ab^T].
\end{align}
Multiplying each side by the pseudo-inverse,
\begin{align}
    ( \Ib + v_t^2 \Ab^T\Ab)^{-1} (\xb_t + v_t^2 \Ab^T \yb) = \xb_0 + \sigma_t  ( \Ib + v_t^2 \Ab^T\Ab)^{-1} (\nb_t + v_t \Ab^T \nb_0).
\end{align}
Following similar arguments, the combined variable $\hat{\xb}_t$ can be written as:
\begin{align}
   \hat{\xb}_t &= ( \Ib + v_t^2 \Ab^T\Ab)^{-0.5} (\xb_t + v_t^2 \Ab^T \yb) \\
   &= ( \Ib + v_t^2 \Ab^T\Ab)^{0.5} \xb_0 + \sigma_t \hat{\nb}_t, \\
   &= \hat{\Ab}_t \xb_0 + \sigma_t \hat{\nb}_t.
\end{align}
with $\hat{\nb}_t \sim \mathcal{N}(0, \Ib)$ and $\hat{\Ab}_t = (\Ib + v_t^2 \Ab^T\Ab)^{0.5}$. Similar arguments to Appendix B can be followed to show that $\hat{\xb}_t$ is the optimal combination for VE-SDE such that,
\begin{align}
    \Expect[\xb_0; \xb_t, \yb] = \Expect[\xb_0; \hat{\xb}_t].
\end{align}
Then, sampling can be performed through:
\begin{align}
     \nabla_{\xb_t} \log q(\xb_t; \yb) = \frac{1}{\sigma_t^2} (\Expect[\xb_0; \hat{\xb}_t] - \xb_t).
\end{align}

Again, following similar arguments to Appendix C, the training of the VE-SDE score-function can be performed using the loss:
\begin{align}
 \mathcal{L}_\text{Bayesian} = \Expect_{q(\hat{\xb}_t, \xb_0)} \Vert s_{\theta}(\hat{\xb}_t, t) - \xb_0 \Vert^2.
\end{align}

\subsection{Image Denoising}
\label{sec:VE-SDE-2}
To demonstrate the VE-SDE variant, we performed an image-denoising experiment. In the denoising settings, $\Ab = \Ib$ and $\sigma_0 = 0.5$. Then,
\begin{align}
    \hat{\Ab}_t &= \sqrt{1 + v_t^2} \Ib, \\
    \hat{\xb}_t &= \sqrt{1 + v_t^2} \xb_0 + \sigma_t \hat{\nb}_t.
\end{align}
For this experiment, we utilized the pre-trained VE-SDE model in Score-SDE library for sampling \cite{scoresde}. The model was trained to estimate the noise from an input noise sample, and the score-function was derived based on this estimate:
\begin{align}
    \xb_t = \xb_0 + \sigma_t \hat{\nb}_t,     s_{\theta}(\xb_t, t) = -\frac{1}{\sigma_t} \hat{\nb}_t.
\end{align}
To match the scale of input samples, we rescaled $\hat{\xb}_t$ via division by $(1 + v_t^2)^{0.5}$. During inference, the score function was estimated using the unconditional score-function $s_\theta(\cdot,t)$:
\begin{align}
    \nabla_{\xb_t} \log p_{\theta}(\xb_t; \yb) &= \frac{1}{\sigma_t^2} \left( \Expect[\xb_0; \xb_t, \yb] - \xb_t \right), \\
    \nabla_{\xb_t} \log p_{\theta}(\xb_t; \yb) &= \frac{1}{\sigma_t^2} \left( \frac{1}{1 + v_t^2}(\xb_t + v_t^2 \yb) + \frac{\sigma_t^2}{1 + v_t^2} s_{\theta}\left( \frac{1}{1 + v_t^2}(\xb_t + v_t^2 \yb), \frac{\sigma_t}{\sqrt{1 + v_t^2}}\right) - \xb_t \right).
\end{align}

\begin{figure}[t]
\centering
		\includegraphics[scale=0.33]{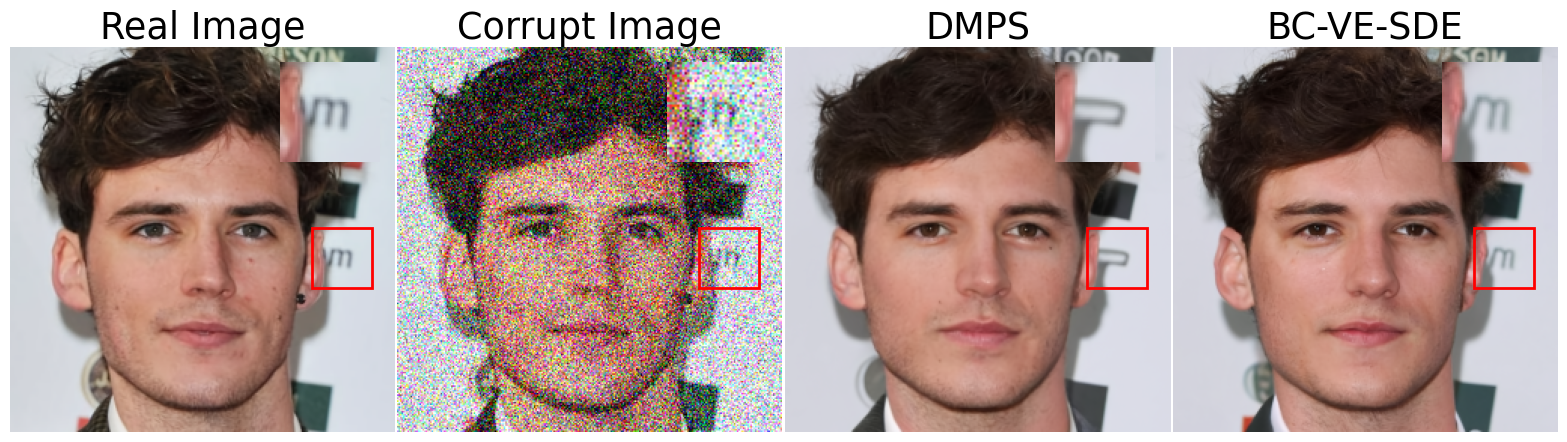}
	\caption{Reconstructed images for the image denoising experiment. The ground-truth (real) image, noise-corrupted measurement, DMPS-recovered image, and BC-VE-SDE-recovered image are displayed.}
	\label{fig:denoising}
\end{figure}

We compared Bayesian conditioned VE-SDE with DMPS \cite{dmps2022diffusion}. To this end, we used the first 480 images in the CelebAHQ dataset. We used reverse diffusion predictor with Langevin corrector for $2000$ timesteps. We reconstructed a single image sample with both methods under an average inference time of $4$ minutes. The quantitative performance metrics for DMPS and BC-VE-SDE are: 26.12$\pm$1.32 versus 26.15$\pm$1.30 pSNR; 73.6$\pm$6.4 versus 73.9$\pm$6.3 SSIM; 38.70 versus 35.74 FID, respectively. While the metrics are, in general, similar, a moderate improvement in FID is achieved by BC-VE-SDE. A representative image is displayed in Fig.~\ref{fig:denoising}. While DMPS fails to recover the letter ``m'' as a detailed feature in the background, the proposed BC-VE-SDE method faithfully recovers it without any additional computational cost.

\end{document}